\documentclass[runningheads]{llncs}
\usepackage{graphicx}
\usepackage[dvipsnames]{xcolor}
\usepackage{mathtools}
\usepackage{amssymb,amsfonts,amsmath} %ams
\usepackage{enumerate,enumitem,tikz,graphicx,mathrsfs,eucal,verbatim, bbm, derivative}
\usepackage{algorithm}
\usepackage{algpseudocode}
\usepackage{caption}
\usepackage{subcaption}
\usepackage{wrapfig}
\usepackage{sidecap}
\usepackage{hyperref}
\usepackage{url}
\usepackage{cite}
\usepackage[misc]{ifsym}

\newcommand{\Sint}{\mathcal{S}_{\textnormal{int}}}
\newcommand{\Sext}{\mathcal{S}_{\textnormal{ext}}}
\newcommand{\sint}{{s^{}_{\textnormal{int}}}}
\newcommand{\sext}{{s^{}_{\textnormal{ext}}}}
\newcommand{\sintp}{s_{\textnormal{int}}'}
\newcommand{\sextp}{s_{\textnormal{ext}}'}
\newcommand{\Zint}{\mathcal{Z}_{\textnormal{int}}}
\newcommand{\Zext}{\mathcal{Z}_{\textnormal{ext}}}
\newcommand{\zint}{{z^{}_{\textnormal{int}}}}
\newcommand{\zext}{{z^{}_{\textnormal{ext}}}}
\newcommand{\zintp}{z_{\textnormal{int}}'}
\newcommand{\zextp}{z_{\textnormal{ext}}'}
\newcommand{\Lint}{\mathcal{L}_{\textnormal{int}}}
\newcommand{\Lext}{\mathcal{L}_{\textnormal{ext}}}
\newcommand{\Lcont}{\mathcal{L}_{\textnormal{cont}}}

\newcommand\blfootnote[1]{%
  \begin{NoHyper}%
  \renewcommand\thefootnote{}\footnote{#1}%
  \addtocounter{footnote}{-1}%
  \end{NoHyper}%
}

\spnewtheorem{assmp}[theorem]{Assumption}{\bfseries}{\itshape}

\begin{document}
\title{Learning Geometric Representations \\ of Objects via Interaction}
%
%\titlerunning{Abbreviated paper title}
% If the paper title is too long for the running head, you can set
% an abbreviated paper title here
%
\author{Alfredo Reichlin (\Letter)*\inst{1} \and
Giovanni Luca Marchetti*\inst{1} \and
Hang Yin\inst{2} \and
Anastasiia Varava\inst{1} \and
Danica Kragic\inst{1}
}
\authorrunning{A. Reichlin, G. L. Marchetti et al.}
% First names are abbreviated in the running head.
% If there are more than two authors, 'et al.' is used.
%
\institute{KTH Royal Institute of Technology, Stockholm, Sweden \\ \email{\{alfrei,glma\}@kth.se} \and
University of Copenhagen, Copenhagen, Denmark
}

\tocauthor{Alfredo~Reichlin,  Giovanni~Luca~Marchetti,  Hang~Yin  Anastasiia~Varava, Danica~Kragic}
\toctitle{Learning Geometric Representations of Objects via Interaction}

\maketitle              % typeset the header of the contribution
\begin{abstract}
We address the problem of learning representations from observations of a scene involving an agent and an external object the agent interacts with. To this end, we propose a representation learning framework extracting the location in physical space of both the agent and the object from unstructured observations of arbitrary nature. Our framework relies on the actions performed by the agent as the only source of supervision, while assuming that the object is displaced by the agent via unknown dynamics. We provide a theoretical foundation and formally prove that an ideal learner is guaranteed to infer an isometric representation, disentangling the agent from the object and correctly extracting their locations. We evaluate empirically our framework on a variety of scenarios, showing that it outperforms vision-based approaches such as a state-of-the-art keypoint extractor. We moreover demonstrate how the extracted representations enable the agent to solve downstream tasks via reinforcement learning in an efficient manner.\blfootnote{*Equal Contribution}

\end{abstract}
\section{Introduction}
A fundamental aspect of intelligent behavior by part of an agent is building rich and structured \emph{representations} of the surrounding world \cite{ha2018world}. Through structure, in fact, a representation potentially leads to  semantic understanding, efficient reasoning and generalization \cite{lake2017building}. However, in a realistic scenario an agent perceives observations of the world that are high-dimensional and unstructured e.g., images. Therefore, the ultimate goal of inferring a representation consists of extracting structure from the observed data \cite{bengio2013representation}. This is challenging and in some instances requires supervision or biases. For example, it is known that \emph{disentangling} factors of variation in data is mathematically impossible in a completely unsupervised way \cite{locatello2019challenging}. In order to extract structure, it is therefore necessary to design methods and paradigms relying on additional information and specific assumptions.

In the context of an agent interacting with the world, a fruitful source of information is provided by the \emph{actions} performed and collected together with the observations. Based on this, several recent works have explored the role of actions in representation learning and proposed methods to extract structure from interaction \cite{kipf2019contrastive, mondal2020group, park2022learning}. The common principle underlying this line of research is encouraging the representation to replicate the effect of actions in a structured space -- a property referred to as \emph{equivariance} \footnote{Alternative terminologies from the literature are \emph{World Model} \cite{kipf2019contrastive} and \emph{Markov Decision Process Homomorphism} \cite{van2020plannable}.}. In particular, it has been shown in \cite{marchetti2022equivariant} that equivariance enables to extract the location of the agent in physical space, resulting in a lossless and geometric representation. The question of how to represent features of the world which are extrinsic to the agent (e.g., objects) has been left open. Such features are dynamic since they change as a consequence of interaction. They are thus challenging to capture in the representation but are essential for understanding and reasoning by part of the agent. 

\begin{figure}[t]
\centering
\includegraphics[width=1.\linewidth]{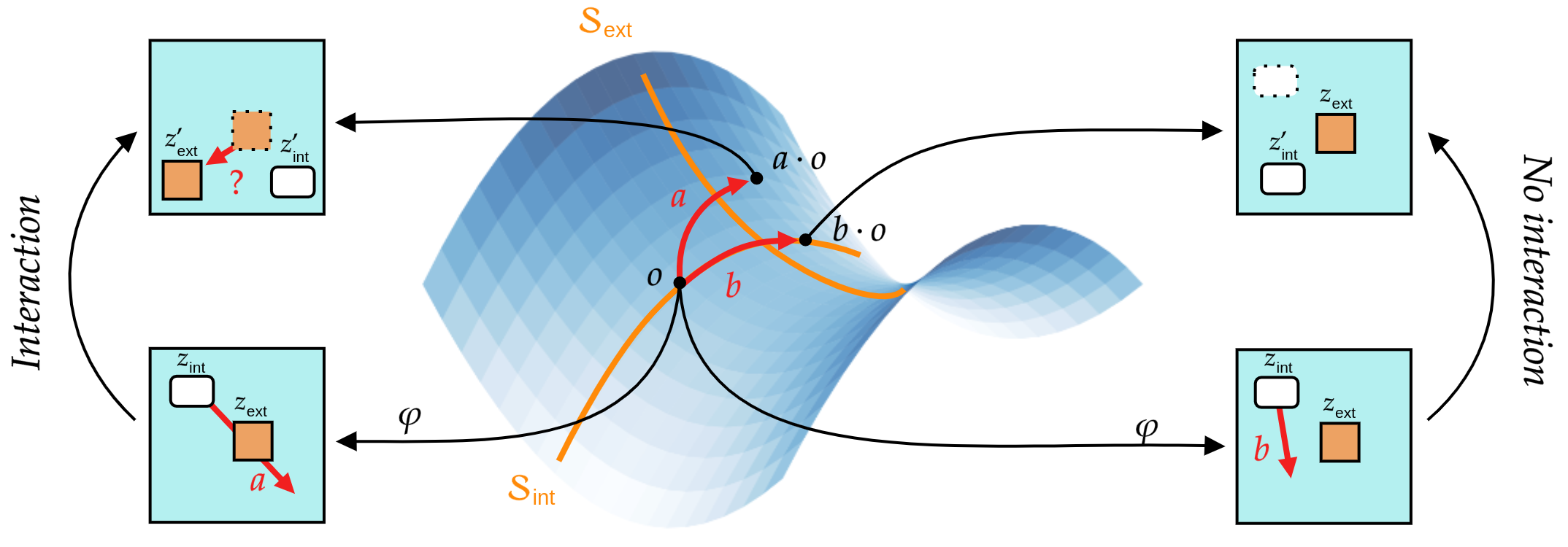}
    \caption{Our framework enables to learn a representation $\varphi$ recovering the geometric and disentangled state of both an agent ($\zint$, white) and an interactable object ($\zext$, brown) from unstructured observations $o$ (e.g., images). The only form of supervision comes from actions $a,b$ performed by the agent, while the transition of the object (question mark) in case of interaction is unknown. In case of no interaction, the object stays invariant. }
\label{firstpage}
\end{figure}

In this work we consider the problem of learning representations of a scene involving an agent and an external rigid object the agent interacts with (see Figure \ref{firstpage}). We aim for a representation disentangling the agent from the object and extracting the locations of both of them in physical space. In order words, we aim for representations that are isometric w.r.t. to the geometry of the world. To this end, we focus on a scenario where the object displaces only when it comes in contact with the agent, which is realistic and practical. We make no additional assumption on the complexity of the interaction: the object is allowed to displace arbitrarily and its dynamics is unknown. Our assumption around the interaction enables to separate the problem of representing the agent -- whose actions are known and available as a supervisory signal -- from the problem of representing the object -- whose displacement is unknown. Following this principle, we design an optimization objective relying on actions as the only form of supervision. This makes the framework general and in principle applicable to observations of arbitrary nature. We moreover provide a formalization of the problem and theoretical grounding for the method. Our core theoretical result guarantees that the representation inferred by an ideal learner recovers isometric representations as desired. We complement the theoretical analysis with an empirical investigation. Results show that our proposed representations outperform in quality of structure a state-of-the-art keypoint extractor and can be leveraged by the agent in order to solve control tasks efficiently by reinforcement learning. In summary, our contributions include: 
\begin{itemize}
\item A representation learning framework extracting representations from observations of a scene involving an agent interacting with an object.  
\item A theoretical result guaranteeing that the above learning framework, when implemented by an ideal learner, infers an isometric representation for data of arbitrary nature.  
\item An empirical investigation of the framework on a variety of environments with comparisons to computer vision approaches (i.e., keypoint extraction) and applications to a control task. 
\end{itemize}

We provide Python code implementing our framework together with all the experiments at the following public repository: \url{https://github.com/reichlin/GeomRepObj}. The repository additionally includes the Appendix of the present work.
\section{Related Work}\label{sec:relwork}
%{\setlength{\parindent}{0cm}
\textbf{Equivariant Representation Learning}. Several recent works have explored the idea of incorporating interactions into representation learning. The common principle is to infer a representation which is equivariant i.e., such that transitions in observations are replicated as transitions in the latent space. One option is to learn the latent transition end-to-end together with the representation \cite{kipf2019contrastive, van2020plannable, watter2015embed}. This approach is however non-interpretable and the resulting representations are not guaranteed to extract any structure. Alternatively, the latent transition can be designed a priori. Linear and affine latent transitions have been considered in \cite{guo2019affine}, \cite{mondal2020group} and \cite{park2022learning} while transitions defined by (the multiplication of) a Lie group have been discussed in \cite{marchetti2022equivariant}, \cite{mondal2022eqr}. As shown in \cite{marchetti2022equivariant}, for static scenarios (i.e., with no interactive external objects) the resulting representations are structured and completely recover the  geometry of the underlying state of the agent. Our framework adheres to this line of research by modelling the latent transitions via the additive Lie group $\mathbb{R}^n$. We however further extend the representation to include external objects. Our framework thus applies to more general scenarios and dynamics while still benefiting from the geometrical guarantees. 
        
%\label{sec:keypoint}
\textbf{Keypoint Extraction}. When observations are images, computer vision offers a spectrum of classical approaches to extract geometric structure. In particular, extracting keypoints enables to identify any object appearing in the observed images. Popular keypoint extractors include classical non-parametric methods \cite{lowe1999object}, \cite{bay2006surf} as well as modern self-supervised learning approaches \cite{kulkarni2019unsupervised}, \cite{gopalakrishnan2020unsupervised}. However, keypoints from an image provide a representation based on the geometry of the field of view or, equivalently, of the pixel plane. This means that the intrinsic three-dimensional geometry of states of objects is not preserved since the representation differs from it by an unknown projective transformation. In specific situations such transformation can still be recovered by processing the extracted keypoints. This is the case when images are in first person view w.r.t. the observer: the keypoints can then be converted into three-dimensional landmarks via methods such as bundle adjustment \cite{triggs1999bundle}, \cite{schonberger2016structure}. Differently from computer vision approaches, our framework is data-agnostic and does not rely on specific priors tied to the nature of observations. It instead extracts representations based on the actions performed by the agent, which is possible due to the dynamical assumptions described in Section \ref{secform}. 

\textbf{Interactive Perception}. The role of interaction in perception has been extensively studied in cognitive sciences and neuroscience \cite{held1963movement, gibson1966senses, noe2004action}. Inspired by those, the field of interactive perception from robotics aims to enhance the understanding of the world by part of an artificial system via interactions \cite{bohg2017interactive}. Applications include active control of cameras \cite{bajcsy1988active} and manipulators \cite{88140} in order to improve the perception of objects \cite{ilonen2014three, 6696808, siftmanipulation}. Our work fits into the program of interactive perception since we crucially rely on performed actions as a self-supervisory signal to learn the representation. We show that the location of objects can be extracted from actions alone, albeit in a particular dynamical setting. Without interaction, this would require strong assumptions and knowledge around the data and the environment as discussed in Section \ref{sec:relwork}. 

\section{Formalism and Assumptions}\label{secform}
In this section we introduce the relevant mathematical formalism together with the assumptions necessary for our framework. We consider the following scenario: an agent navigates in a Euclidean space and interacts in an unknown way with an external object. This means that the space of states $\mathcal{S}$ is decomposed as 
\begin{equation}
\mathcal{S}=\Sint \times \Sext
\end{equation}
where $\Sint$ is the space of states of the agent (\emph{internal} states) and $\Sext$ is the space of states of the object (\emph{external} states). We identify both the agent and the object with their location in the ambient space, meaning that $\Sint \subseteq \mathbb{R}^n \supseteq \Sext$, where $n$ is the ambient dimension. The actions that the agent performs are displacements of its state i.e., the space of actions consists of translations $\mathcal{A} = \mathbb{R}^n$. In our formalism we thus abstract objects as material points for simplicity of the theoretical analysis. The practical extension to volumetric objects together with their orientation is discussed in Section \ref{sec:stocha} while the extension of agent's actions to arbitrary Lie groups is briefly discussed in Section \ref{sec:concfut}.

Our first assumption is that the agent can reach any position from any other via a sequence of actions. This translates in the following connectivity condition:

\begin{assmp} \label{ass:conn}
\textnormal{(Connectedness)} The space $\Sint$ is connected and open.
\end{assmp}
When the agent performs an action $a \in \mathcal{A}$ the state $s = (\sint, \sext)$ transitions into a novel one denoted by $a \cdot s = (\sintp, \sextp)$. Since the actions displace the agent, the internal state gets translated as $\sintp = \sint + a$.\footnote{Whenever we write $a \cdot s$ we implicitly assume that the action is valid i.e., that $\sint + a \in \Sint$.} However, the law governing the transition of the object $\sextp = T(s, a)$ is assumed to be unknown and can be arbitrarily complex and stochastic. We stick to deterministic transitions for simplicity of explanation. Crucially, the agent does not have access to the ground-truth state $s$. Instead it perceives unstructured and potentially high-dimensional observations $o \in \mathcal{O}$ (e.g., images) via an unknown emission map $\omega: \ \mathcal{S} \rightarrow \mathcal{O}$. We assume that $\omega$ is injective so that actions induce deterministic transitions of observations, which we denote as $o' = a \cdot o$. This assumption is equivalent to total observability of the scenario and again simplifies the forthcoming discussions by avoiding the need to model stochasticity in $\mathcal{O}$. 

The fundamental assumption of this work is that the dynamics of the external object revolves around \emph{contact} i.e., the object does not displace unless it is touched by the agent. This is natural and often satisfied in practice. In order to formalize it, note that when the agent in state $\sint$ performs an action $a \in \mathcal{A}$ we can imagine it moving along the open segment $\lfloor \sint, \ \sint + a \rfloor = \{ \sint 
+ ta \}_{0 < t < 1}$. Our assumption then translates into (see Figure \ref{firstpage} for a graphical depiction):

\begin{assmp}\label{ass:cont}
\textnormal{(Interaction Occurs at Contact)} For all agent states $\sint \in S$ and actions $a \in \mathcal{A}$ it holds that $\sextp = \sext$ if and only if $\sext \not \in \lfloor \sint, \ \sint + a \rfloor $. 
\end{assmp}
As such, the dynamics of the external object can be summarized as follows:
\begin{equation}\label{eq:trandyn}
    \sextp = \begin{cases} \sext& \text{if } \ \sext \not \in \lfloor \sint, \ \sint + a \rfloor, \\ T(s, a) & \text{otherwise.} \end{cases}
\end{equation}

Finally, we need to assume that interaction is possible for every state of the object i.e., the latter has to be always reachable by the agent. This is formalized via the following inclusion: 

\begin{assmp}\label{ass:rea}
\textnormal{(Reachability)} It holds that $\Sext \subseteq \Sint$. 
\end{assmp}

\section{Method}
\subsection{Representations and Equivariance}\label{sec:repequi}
We now outline the inference problem addressed in the present work. Given the setting introduced in Section \ref{secform}, the overall goal is to infer a \emph{representation} of observations $\varphi: \ \mathcal{O} \rightarrow \mathcal{Z} = \Zint \times \Zext $, where $\Zint = \Zext  = \mathbb{R}^n$. Ideally $\varphi$ recovers the underlying inaccessible state in $\mathcal{S} \subseteq \mathcal{Z}$ and disentangles $\Sint$ from $\Sext$. In order to achieve this, our central idea is to split the problem of representing the agent and the object. Since the actions of the agent are available, $\zint \in \Zint$ can be inferred geometrically by existing representation learning methods. The representation of the object $\zext \in \Zext$ can then be inferred based on the one of the agent by exploiting the relation between the dynamics of the two (Equation \ref{eq:trandyn}). In order to represent the agent, we consider the fundamental concept of (translational) \emph{equivariance}:
\begin{definition}\label{defequiv}
    The representation $\varphi$ is said to be \emph{equivariant} (on internal states) if for all $a \in \mathcal{A}$ and $o \in \mathcal{O}$ it holds that $\zintp = \zint + a$ where  $(\zint, \zext) = \varphi(o)$ and $(\zintp, \zextp) = \varphi(a \cdot o)$. 
\end{definition}

We remark that Definition \ref{defequiv} refers to internal states only, making our terminology around equivariance unconventional. As observed in previous work \cite{marchetti2022equivariant}, equivariance guarantees a faithful representation of internal states. Indeed if $\varphi$ is equivariant then $\zint$ differs from $\sint$ by a constant vector. This means that the representation of internal states is a translation of ground-truth ones and as such is lossless (i.e., bijective) and isometrically recovers the geometry of $\Sint$. 

The above principle can be leveraged in order to learn a representation of external states with the same benefits as the representation of internal ones. Since the external object displaces only when it comes in contact with the agent (Assumption \ref{ass:cont}), the intuition is that $\zext$ can be inferred by aligning it with $\zint$. The following theoretical result formalizes the possibility of learning such representations and traces the foundation of our learning framework. 

\begin{theorem}\label{thminter}
Suppose that the representation $\varphi: \ \mathcal{O} \rightarrow \mathcal{Z}$ satisfies: 
\begin{enumerate}
 \item $\varphi$ is equivariant (Definition \ref{defequiv}),
 \item $\varphi$ is injective, 
 \item  for all $o \in \mathcal{O}$ and $a \in \mathcal{A}$ it holds that either $\zextp  = \zext$ or $\zext \in \lfloor \zint, \zint + a \rfloor $ where $(\zint, \zext) = \varphi(o)$ and $(\zintp, \zextp) = \varphi(a \cdot o)$. 
 \end{enumerate}
 Then $\varphi \circ \omega$ is a translation i.e., there is a constant vector $h \in \mathbb{R}^n$ such that for all $s \in \mathcal{S}$ it holds that $\varphi(\omega(s)) = s + h$. In particular, $\varphi \circ \omega$ is an isometry w.r.t. the Euclidean metric on both $\mathcal{S}$ and $\mathcal{Z}$.
\end{theorem}

We refer to the Appendix for a proof. Theorem \ref{thminter} states that if the conditions $1. - 3.$ are satisfied (together with the assumptions stated in Section \ref{secform}) then the representation recovers the inaccessible state up to a translation and thus isometrically preserves the geometry of the environment. All the conditions from Theorem \ref{thminter} refer to properties of $\varphi$ depending on observations and the effect of actions on them, which are accessible in practice. The goal of the forthcoming section is to describe how these conditions can be enforced on $\varphi$ by optimizing a system of losses. 

\subsection{Learning the Representation}

In this section we describe a viable implementation of a representation learning framework adhering to the conditions of Theorem \ref{thminter}.  We model the representation learner $\varphi = (\varphi_{\textnormal{int}} , \varphi_{\textnormal{ext}})$ as two parameterized functions $\varphi_{\textnormal{int}} : \ \mathcal{O} \rightarrow \Zint$,  $\varphi_{\textnormal{ext}} : \ \mathcal{O} \rightarrow \Zext$ e.g., two deep neural network models. In order to train the models, we assume that the dataset $\mathcal{D}$ consists of transitions observed by the agent in the form of $\mathcal{D} = \{ (o, a, o' = a \cdot o) \} \subseteq \mathcal{O} \times \mathcal{A} \times \mathcal{O}$. Such data can be collected by the agent autonomously exploring its environment and randomly interacting with the external object. This implies that the only form of supervision required consists of the actions performed by the agent together with their effect on the observations. 

First, we propose to enforce equivariance, condition $1$ from Theorem \ref{thminter}, by minimizing the loss:  
\begin{equation}\label{eqlint}
\Lint(o, a, o') =  d(\zintp, \zint + a ) 
\end{equation}
where $d$ is a measure of similarity on $\Zint = \mathbb{R}^n$ and the notation is in accordance with Definition \ref{defequiv}. Typically $d$ is chosen as the squared Euclidean distance as described in previous work \cite{mondal2020group, kipf2019contrastive}.

Next, we focus on the representation of the external object. As stated before, the dataset consists of transitions either with or without interaction. When an interaction occurs, $\zext$ should belong to the segment $\lfloor \zint, \zint + a \rfloor$. When it doesn't, the representation should be invariant i.e., $\zext = \zextp$. These two cases are outlined in condition 2 of Theorem \ref{thminter} and can be enforced via the following losses:
\begin{equation}\label{eqposneg}
\mathcal{L}_-(o, a, o') = d(\zext, \zextp) \hspace{1cm} \mathcal{L}_+(o, a, o') = d(\zext, \lfloor \zint, \zint + a \rfloor ).
\end{equation}

The distance involved in $\mathcal{L}_+$ represents a point-to-set metric and is typically set as $d(z, E) = \inf_{x \in E }d(z,x)$. The latter has a simple explicit expression in the case $E$ is a segment.

\begin{figure}[h!] %{0.5\textwidth}
  \begin{center}
    \includegraphics[width=.33\textwidth]{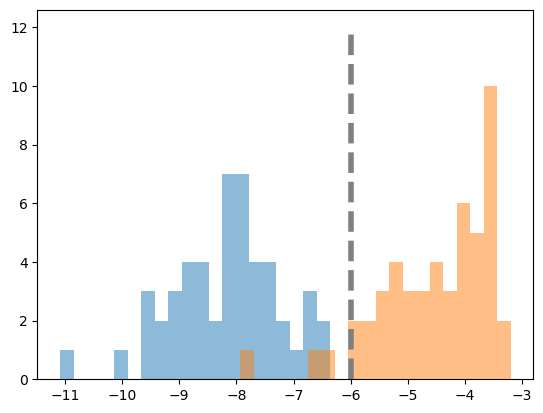}
  \end{center}
    \begin{picture}(0,0)
\put(150,10){$\log d_\mathcal{W}(w,w')$}
\end{picture}
  \caption{Histograms of the log-distances in $\mathcal{W}$. Colors indicate whether interaction occurs (orange) or not (blue). The dotted line represents the threshold from Otsu's algorithm. }\label{histo}
\end{figure}
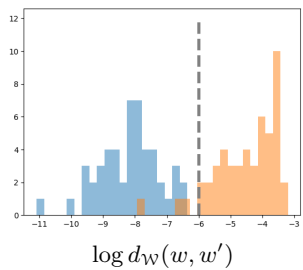

However, the data contains no information on whether interaction occurs or not. It is, therefore, necessary to design a procedure determining when to optimize $\mathcal{L}_+$ and  $\mathcal{L}_-$. To this end, we propose to train a parallel model $\varphi_{\textnormal{cont}}: \mathcal{O} \rightarrow \mathcal{W}$ with latent \emph{contrastive representation} $\mathcal{W}$ (potentially different from $\mathcal{Z}$). This is trained to attract $w=\varphi_{\textnormal{cont}}(o)$ to $w'=\varphi_{\textnormal{cont}}(o')$ while forcing injectivity of $\varphi$ (condition 2 from Theorem \ref{thminter}). To this end, we stick to the popular \emph{InfoNCE} loss from contrastive learning literature \cite{chen2020simple}:

\begin{equation}\label{infonce}
\Lcont(o, o') = 
d_\mathcal{W}(w,w') + \log \mathbb{E}_{o''} \left[ e^{-d_\mathcal{W}(w', w'') - d(\zintp, \zintp')} \right]
\end{equation}

where $o''$ is marginalized from $\mathcal{D}$. The second summand of Equation \ref{infonce} encourages the joint encodings $(\zint, w)$ to spread apart and thus encourages $\varphi$ to be injective. Since subsequent observations where interaction does not occur share the same external state, these will lie closer in $\mathcal{W}$ than the ones where interaction does not occur. This enables to exploit distances in $\mathcal{W}$ in order to choose whether to optimize $\mathcal{L}_-$ or $\mathcal{L}_+$. We propose to partition (the given batch of) the dataset in two disjoint classes $\mathcal{D} = C_- \sqcup C_+$ by applying a natural thresholding algorithm to the quantities $d_\mathcal{W}(w,w')$. This can be achieved via one-dimensional $2$-means clustering, which is equivalent to Otsu's algorithm \cite{otsu1979threshold} (see Figure \ref{histo} for an illustration). We then optimize:
\begin{equation}
\Lext(o, a, o') = \begin{cases}\mathcal{L}_-(o, a, o') &   \textnormal{if }  (o,a,o') \in C_-,\\
\mathcal{L}_+(o, a, o')  &   \textnormal{if }  (o,a,o') \in C_+.
\end{cases}
\end{equation}

In summary, the total loss minimized by the models $(\varphi_{\textnormal{int}}, \varphi_{\textnormal{ext}}, \varphi_{\textnormal{cont}})$ w.r.t. the respective parameters is (see the pseudocode included in the Appendix):
\begin{equation}
\mathcal{L} = \mathbb{E}_{(o,a,o') \sim \mathcal{D}}[\Lint(o,a,o') + \Lext(o,a,o') + \Lcont(o,o') ].
\end{equation}

\subsection{Incorporating Volumes of Objects}\label{sec:stocha}
So far we have abstracted the external object as a point in Euclidean space. However, the object typically manifests with a body and thus occupies a volume. Interaction and consequent displacement (Assumption \ref{ass:rea}) occur when the agent comes in contact with the boundary of the object's body. The representation thus needs to take volumetric features into account in order to faithfully extract the geometry of states.

In order to incorporate volumetric objects into our framework we propose to rely on
\emph{stochastic} outputs i.e., to design $\zext$ as a probability density over $\Zext$ representing (a fuzzy approximation of) the body of the object. More concretely, the output of $\varphi_\textnormal{ext}$ consists of (parameters of) a Gaussian distribution whose covariance matrix represents the inertia ellipsoid of the object i.e., the ellipsoidal approximation of its shape. By diagonalizing the covariance matrix via an orthonormal frame, the orientation of the object can be extracted in the form of a rotation matrix in $\textrm{SO}(n)$. The losses of our model are naturally adapted to the stochastic setting as follows. The distance $d$ appearing in Equation \ref{eqposneg} is replaced with Kullback-Leibler divergence. The latter has an explicit simple expression for Gaussian densities which allows to compute $\mathcal{L}_-$ directly. In order to compute $\mathcal{L}_+$ we rely on a Monte Carlo approximation, meaning that we sample a point uniformly from the interval and set $\mathcal{L}^+$ as the negative log-likelihood of the point w.r.t. the density defining $\zext$. 
\section{Experiments}
We empirically investigate the performance of our framework in correctly identifying the position of an agent and of an interactive object. The overall goal of the experimental evaluation is to show that our representation is capable of extracting the geometry of states without relying on any prior knowledge of observations e.g., depth information. All the scenarios are normalized so that states lie in the unit cube. Observations are RGB images of resolution $100 \times 100$ in all the cases considered. We implement each of $\varphi_{\textnormal{int}}$, $\varphi_{\textnormal{ext}}$ and $\varphi_{\textnormal{cont}}$ as a ResNet-18 \cite{he2016deep} and train them for $100$ epochs via the Adam optimizer with learning rate $0.001$ and batch-size $128$. We compare our framework with two baselines:
\begin{itemize}
    \item \textit{Transporter Network} \cite{kulkarni2019unsupervised}: a vision-based state-of-the-art unsupervised keypoint extractor. The approach heavily relies on image manipulation in order to infer regions of the pixel plane that are persistent between pairs of images. We train the model in order to extract two (normalized) keypoints representing $\zint$ and $\zext$ respectively. 
    
    \item \textit{Variational AutoEncoder} (VAE) \cite{kingma2013auto, rezende2014stochastic}: a popular representation learner with a standard Gaussian prior on its latent space. We impose the prior on $\Zext$ only, while $\varphi_{\textnormal{int}}$ is still trained via the equivariance loss (Equation \ref{eqlint}). The decoder takes the joint latent space $\mathcal{Z}$ in input. We set $\textnormal{dim}(\Zext)=32$. This makes the representations disentangled, so that $\zint$ and $\zext$ are well-defined. The resulting representation of the object is generic and is not designed to extract any specific structure from observations. 
    \end{itemize}

In order to evaluate the preservation of geometry we rely on the following evaluation metric $ \mathcal{L}_{\textnormal{test}}$. Given a trained representation $\varphi: \mathcal{O} \rightarrow \mathcal{Z}$ and a test set $\mathcal{D}_{\textnormal{test}}$ of observations with known ground-truth states, we define: 
\begin{equation}\label{eq:evmet}
    \mathcal{L}_{\textnormal{test}} = \mathbb{E}_{o \sim \mathcal{D}_{\textnormal{test}}}\left[ \ d( \zint - \zext , \sint - \sext) \ \right] 
\end{equation}
where $d$ is the squared Euclidean distance. Since both our framework and (the encoder of) VAE have stochastic outputs (see Section \ref{sec:stocha}), we set $\zext$ as the mean of the corresponding Gaussian distribution. Equation \ref{eq:evmet} measures the quality of preservation of the relative position between the agent and the object by part of the representation. When $\mathcal{L}_{\textnormal{test}} =0$, $\varphi$ is an isometry (w.r.t. the Euclidean metric) and thus recovers the geometry of states. The translational invariance of $\mathcal{L}_{\textnormal{test}}$ makes the comparison agnostic to any reference frame eventually inferred by the given learner.

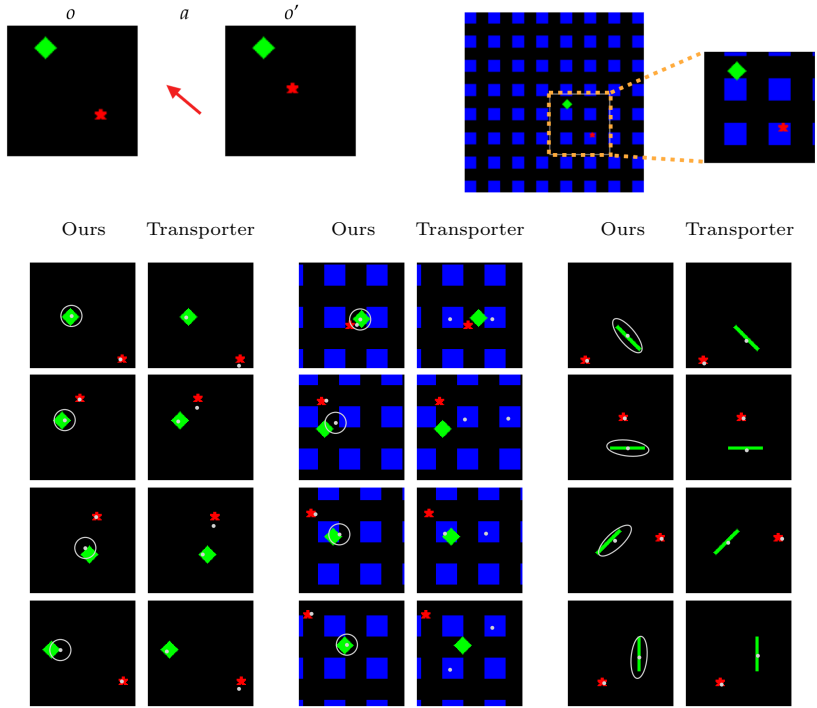
\begin{figure}[th!]
\centering
\begin{subfigure}[b]{.4\linewidth}
        \centering
         \includegraphics[width=\linewidth]{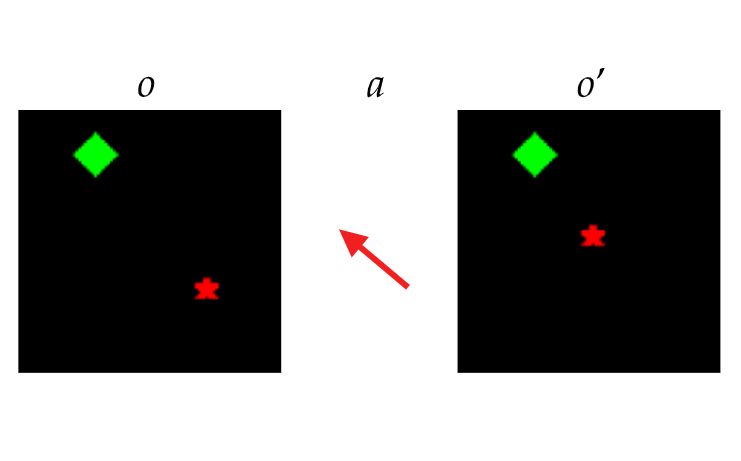}
\end{subfigure}
\hspace{3em}
\begin{subfigure}[b]{.4\linewidth}
        \centering
      \includegraphics[width=\linewidth]{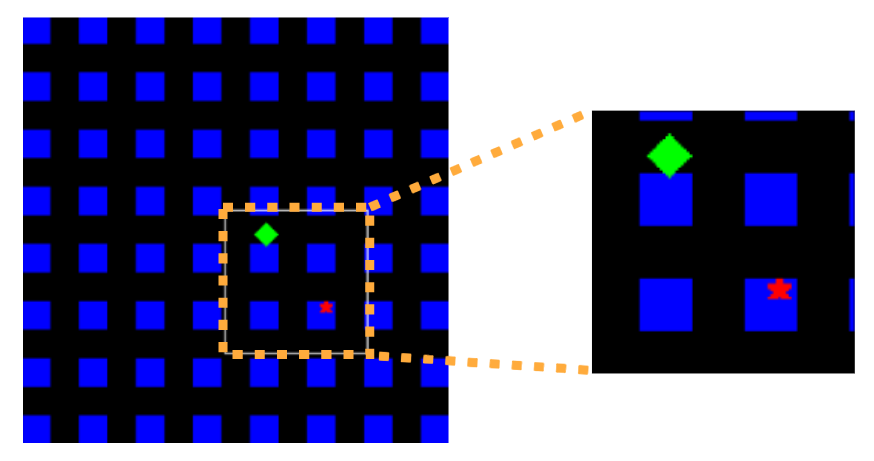}
\end{subfigure}

\hfill \break

\includegraphics[width=.12\linewidth]{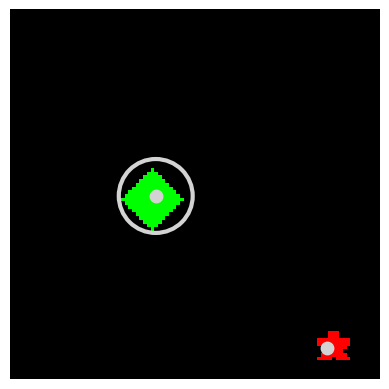}
\includegraphics[width=.12\linewidth]{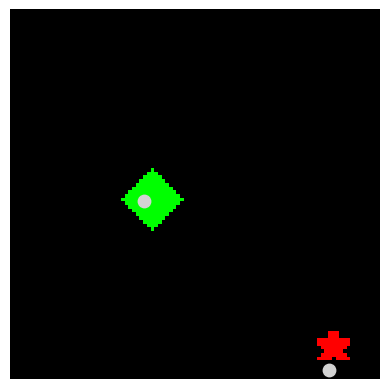}
\hspace{1em}
\includegraphics[width=.12\linewidth]{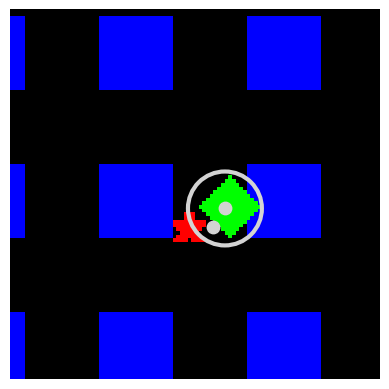}
\includegraphics[width=.12\linewidth]{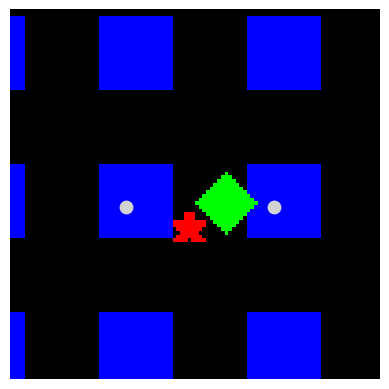}
\hspace{1em}
\includegraphics[width=.12\linewidth]{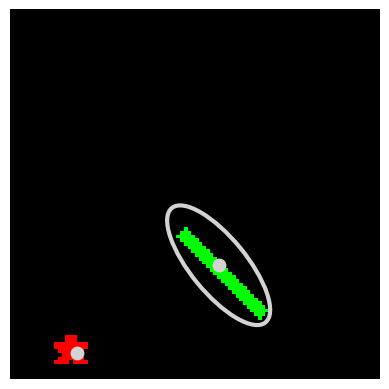}
\includegraphics[width=.12\linewidth]{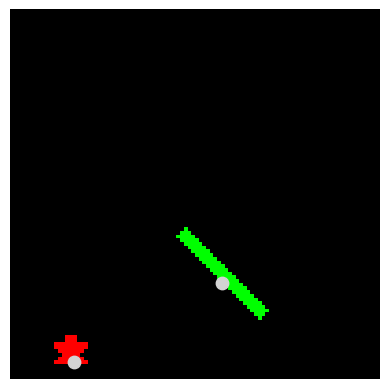}

\includegraphics[width=.12\linewidth]{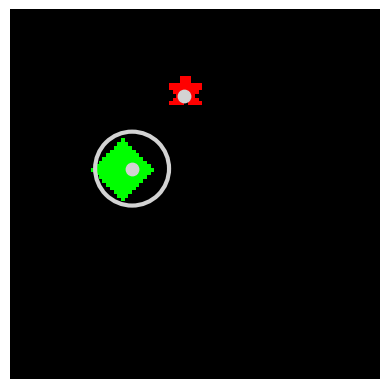}
\includegraphics[width=.12\linewidth]{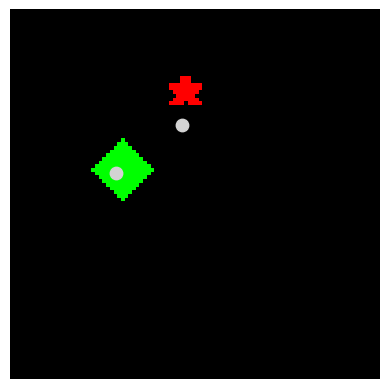}
\hspace{1em}
\includegraphics[width=.12\linewidth]{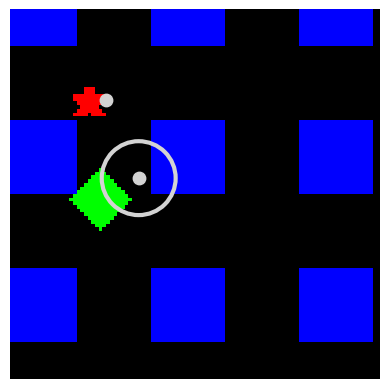}
\includegraphics[width=.12\linewidth]{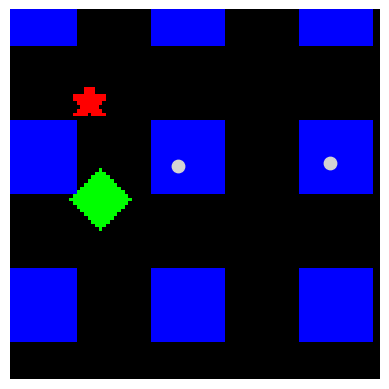}
\hspace{1em}
\includegraphics[width=.12\linewidth]{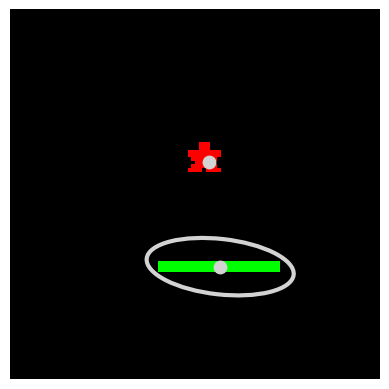}
\includegraphics[width=.12\linewidth]{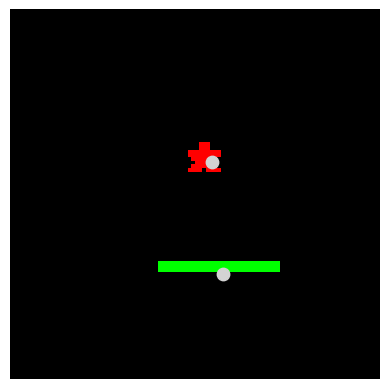}

\includegraphics[width=.12\linewidth]{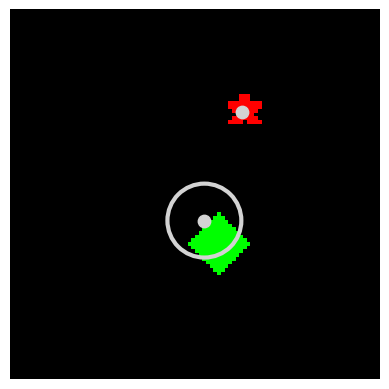}
\includegraphics[width=.12\linewidth]{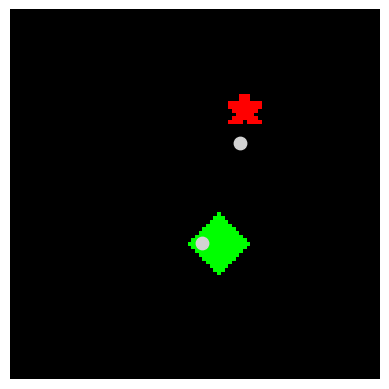}
\hspace{1em}
\includegraphics[width=.12\linewidth]{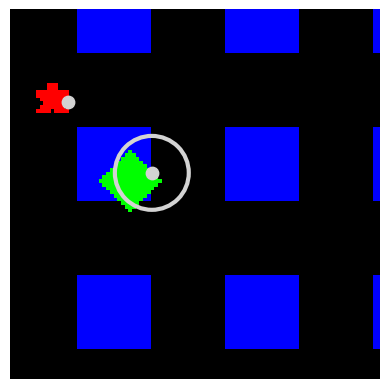}
\includegraphics[width=.12\linewidth]{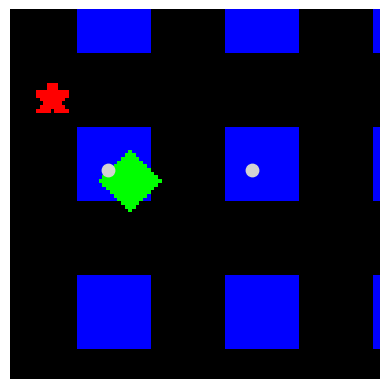}
\hspace{1em}
\includegraphics[width=.12\linewidth]{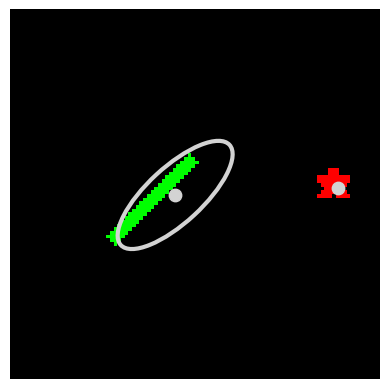}
\includegraphics[width=.12\linewidth]{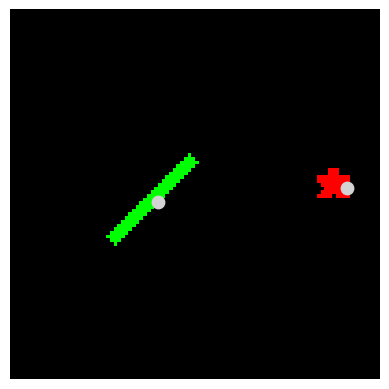}

\includegraphics[width=.12\linewidth]{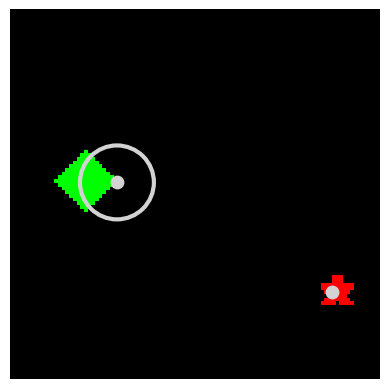}
\includegraphics[width=.12\linewidth]{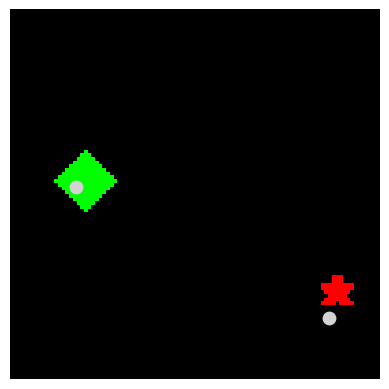}
\hspace{1em}
\includegraphics[width=.12\linewidth]{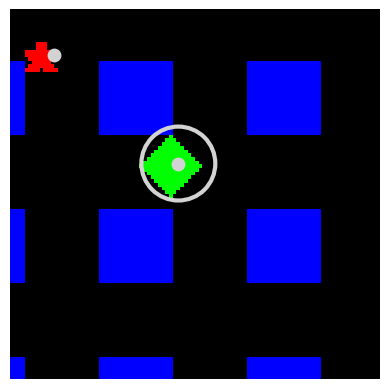}
\includegraphics[width=.12\linewidth]{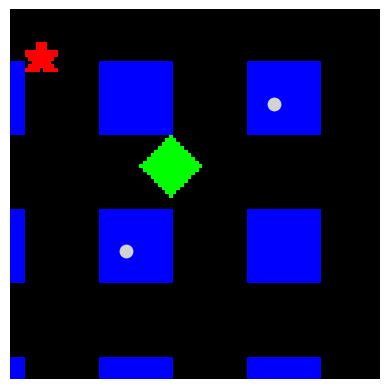}
\hspace{1em}
\includegraphics[width=.12\linewidth]{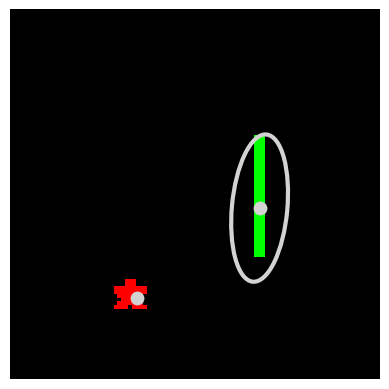}
\includegraphics[width=.12\linewidth]{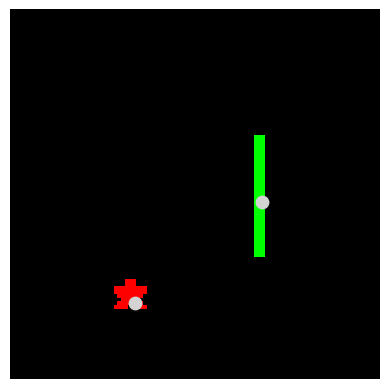}

\begin{picture}(0,0)
    \put(-132,190){{\scriptsize  	 Ours}}
    \put(-100,190){{\scriptsize	 Transporter}}
        \put(-30,190){{\scriptsize  	 Ours}}
    \put(2,190){{\scriptsize	 Transporter}}
        \put(72,190){{\scriptsize  	 Ours}}
    \put(104,190){{\scriptsize	 Transporter}}
\end{picture}

\caption{\textbf{Top:} Visualization of the dataset from the Sprites experiment. On the left, an example of a datapoint $(o, a, o') \in \mathcal{D}$. On the right, an example of an observation from the second version of the dataset where a dynamic background is added as a visual distractor. \textbf{Bottom:} Comparison of $\zint$, $\zext$ (gray dots, with the ellipse representing the learned std) extracted via our model and the Transporter network on the three versions of the Sprites dataset: vanilla version (left), with dynamic background (middle) and with anisotropic object (right). }
  \label{fig:2d_env}
\end{figure}

\subsection{Sprites}

For the first experiment we procedurally generate images of two sprites (the agent and the object) moving on a black background (see Figure \ref{fig:2d_env}, top-left). Between images, the agent (red figure) moves according to a known action. If the agent comes in contact with the object (green diamond) during the execution of the action (see Assumption \ref{ass:cont}) the object is randomly displaced on the next image. In other words, the object's transition function $T(s,a)$ is stochastic with a uniform distribution. Such a completely stochastic dynamics highlights the independence of the displacement of the agent w.r.t. the one of the object. We generate the following two additional versions of the dataset: 
\begin{itemize}
\item A version with \emph{dynamic background}. Images are now overlaid on top of a nine-times larger second image (blue squares in Figure \ref{fig:2d_env}, top-right). The field of view and thus the background moves together with the agent. The background behaves as a visual distractor and makes it challenging to extract structure (e.g., keypoints) via computer vision. 
\item A version with \emph{anisotropic object}. The latter is now a rectangle with one significantly longer side. Besides translating, the object rotates as well when interaction occurs. The goal here is showcasing the ability of our model in inferring the orientation of the object as described in Section \ref{sec:stocha}. 
\end{itemize}
\begin{figure}[h!]
\centering
\includegraphics[width=.7\linewidth]{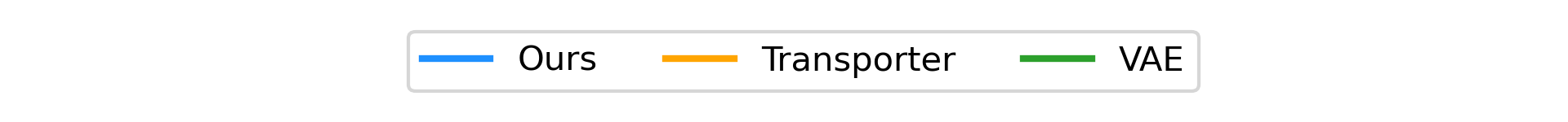}
\begin{subfigure}[b]{.4\linewidth}
        \centering
         \includegraphics[width=\linewidth]{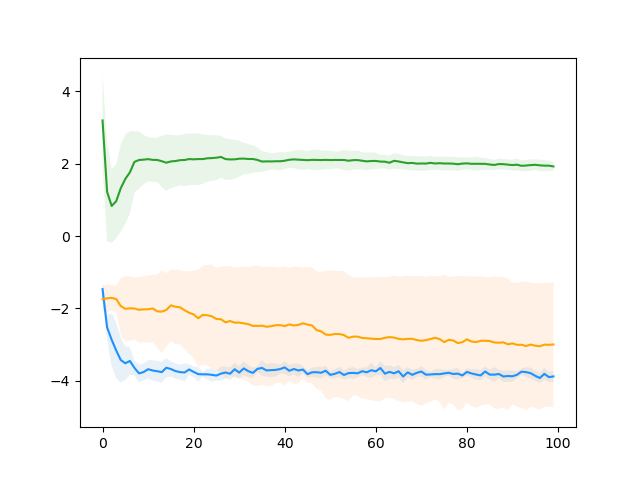}
\end{subfigure}
\hspace{2em}
\begin{subfigure}[b]{.4\linewidth}
        \centering
      \includegraphics[width=\linewidth]{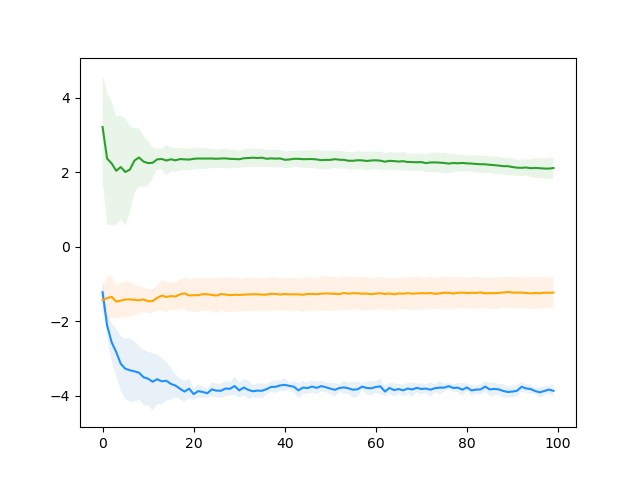}
\end{subfigure}

\begin{picture}(0,0)
\put(-92,8){\scriptsize Epochs}
\put(-150,50){\rotatebox{90}{\scriptsize $\log \mathcal{L}_{\textnormal{test}}$}}
\put(72,8){\scriptsize Epochs}
\put(13,50){\rotatebox{90}{\scriptsize $\log \mathcal{L}_{\textnormal{test}}$}}
\end{picture}
\caption{Log-scale plots of the evaluation metric (Equation \ref{eq:evmet}) as the training progresses for the Sprite experiment. The curves display mean and std (for 10 experimental runs). {\textbf{Left}:} vanilla version of the dataset. \textbf{Right:} version with a dynamic background.}
\label{2d_exp}
\end{figure}
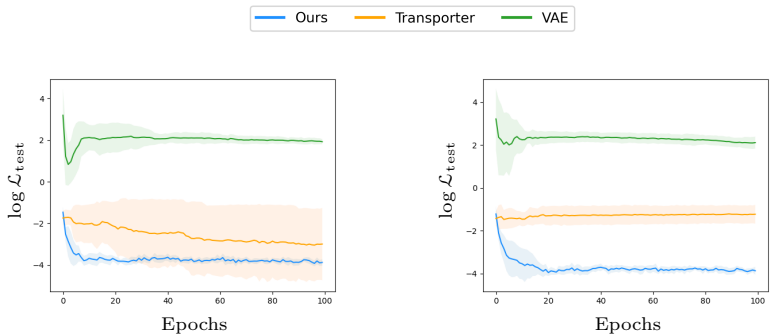

Figure \ref{2d_exp} displays the analytic comparison of the performances between our model and the baselines in terms of the evaluation metric (Equation \ref{eq:evmet}). The plot is in log-scale for visualization purposes. Moreover, Figure \ref{fig:2d_env} (bottom) reports a qualitative comparison between our model and the Transporter network. As can be seen, for the simpler version of the experiment (plot on the left) both our model and the Transporter network successfully achieve low error and recover the geometry of both the agent and the object. Note that the Transporter network converges slowly and with high variance (Figure \ref{2d_exp}, left). This is probably due to the presence of a decoder in its architecture. Our framework instead involves losses designed directly in the latent space, avoiding an additional model to decode observations. As expected, VAE achieves significantly worse performances because of the lack of structure in its representation. As can be seen from Figure \ref{fig:2d_env} (bottom-right), when the object is anisotropic our model correctly infers its orientation by encoding it into the covariance of the learned Gaussian distribution. The Transporter network instead places a keypoint on the barycenter of the object and is therefore unable to recover the orientation.  

For the more challenging version of the experiment with dynamic background, the transporter is not able to extract the expected keypoints. As can be seen from Figure \ref{fig:2d_env} (bottom-middle), the distracting background causes the model to focus on regions of the image not corresponding to the agent and the object. This is reflected by a significantly higher error (and variance) w.r.t. our framework (Figure \ref{2d_exp}, right). The latter still infers the correct representation and preserves geometry. This empirically confirms that our model is robust to visual distractors since it does not rely on any data-specific feature or structure.

\subsection{Soccer}
\begin{figure}[th!]
\centering
\begin{subfigure}[b]{.45\linewidth}
        \centering
 \includegraphics[width=.67\linewidth]{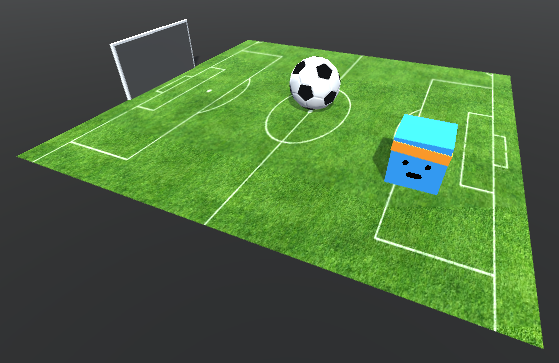}

\hfill \break

 \includegraphics[width=.72\linewidth]{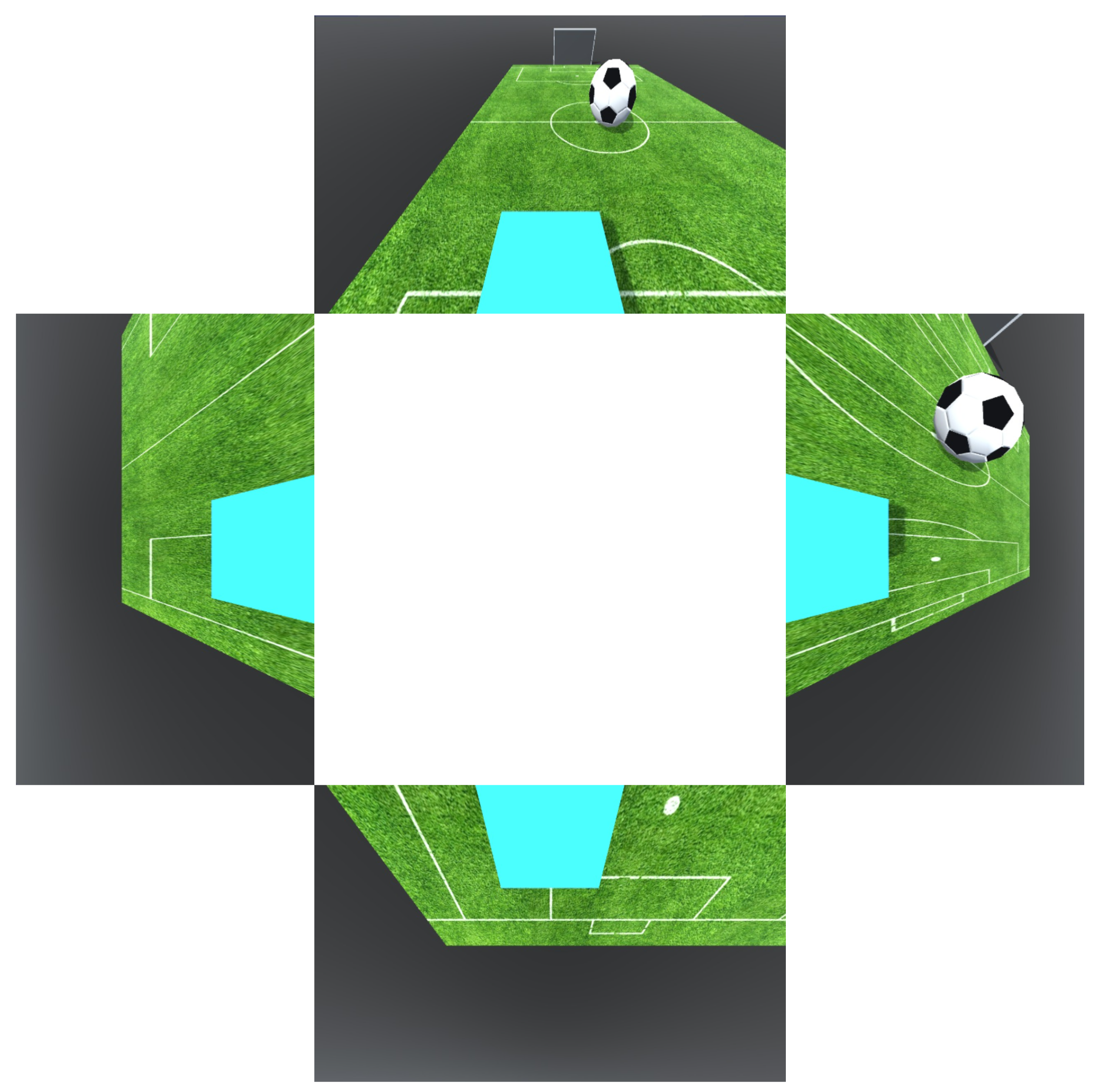}

\end{subfigure}
\begin{subfigure}[b]{.5\linewidth}
\centering

\includegraphics[width=.27\linewidth]{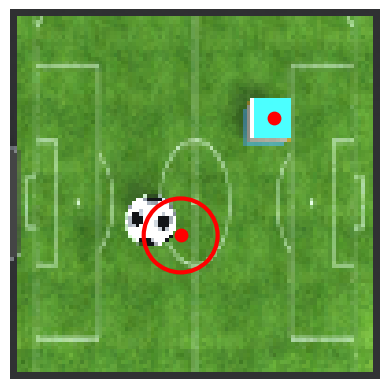}
\includegraphics[width=.27\linewidth]{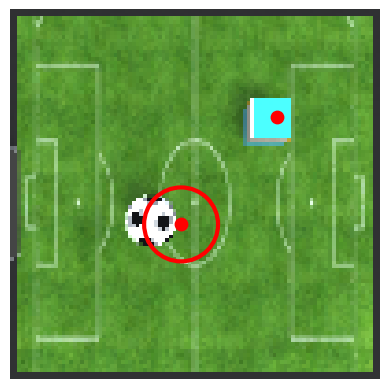}
\includegraphics[width=.27\linewidth]{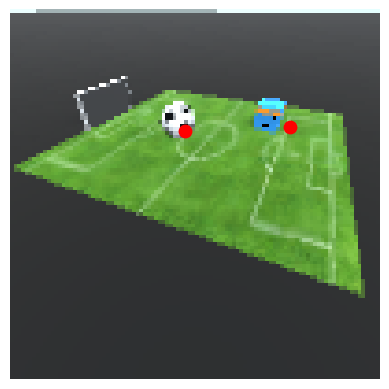}

\includegraphics[width=.27\linewidth]{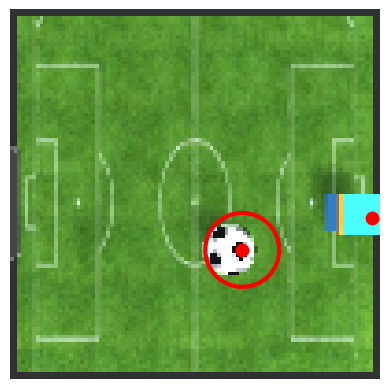}
\includegraphics[width=.27\linewidth]{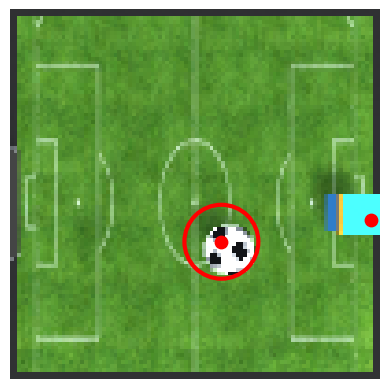}
\includegraphics[width=.27\linewidth]{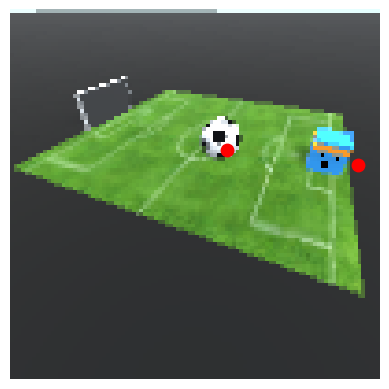}

\includegraphics[width=.27\linewidth]{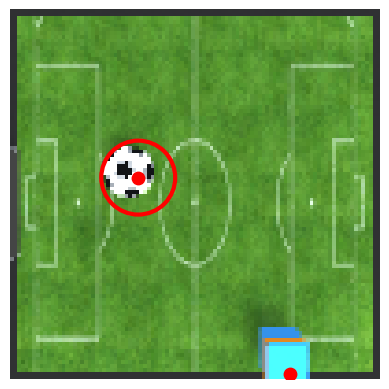}
\includegraphics[width=.27\linewidth]{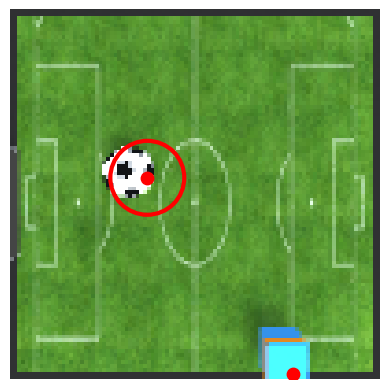}
\includegraphics[width=.27\linewidth]{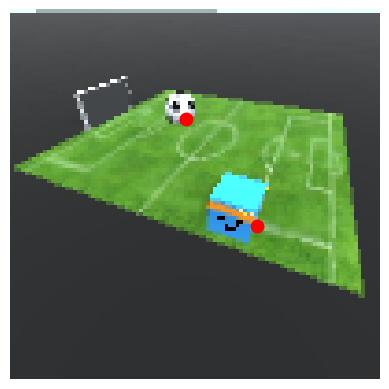}

\includegraphics[width=.27\linewidth]{images/3_soc_ours.png}
\includegraphics[width=.27\linewidth]{images/3_soc_pov_ours.png}
\includegraphics[width=.27\linewidth]{images/3_soc_trans.png}

\end{subfigure}

\begin{picture}(0,0)
    \put(23,220){{\scriptsize Ours}}
    \put(13,210){{\scriptsize  	 (Third P.)}}
    \put(72,220){{\scriptsize  	 Ours}}
    \put(65,210){{\scriptsize  	 (First P.)}}
    \put(110,210){{\scriptsize	 Transporter}}
    \put(-160,44){\rotatebox{90}{\scriptsize First Person}}
    \put(-160,157){\rotatebox{90}{\scriptsize Third Person}}
\end{picture}
\caption{\textbf{Left}: an example of the two types of observations (third and first person respectively) from the Soccer experiment. \textbf{Right}: visual comparison of $\zint$, $\zext$ (red dots) extracted via our model (from third-person view and first-person view) and the Transporter network. For our model, we overlap the representation to a view of the scene from the top instead of the original observation. }
\label{3d_plot}
\end{figure}

For the second experiment we test our framework on an environment consisting of an agent on a soccer field colliding with a ball (see Figure \ref{3d_plot}, left). The scene is generated and rendered via the Unity engine. The physics of the ball is simulated realistically: in case of contact, rolling takes gravity and friction into account. Note that even though the scene is generated via three-dimensional rendering, the (inaccessible) state space is still two-dimensional since the agent navigates on the field. We generate two datasets of $10000$ triples $(o,a,o'=a\cdot o)$ with observations of different nature. The first one consists of views in third-person perspective from a fixed external camera. In the second one, observations are four views in first-person perspective from four cameras attached on top of the agent and pointing in the 4 cardinal directions. We refer to Figure \ref{3d_plot} (left) for a visualization of the two types of observations. In Figure \ref{3d_plot} (right), we report visualizations of the learned representations. The extracted representation of our proposed method depends solely on the geometry of the problem at hand rather than the nature of the observation. The learned representation is thus identical when learned from the third-person dataset or the first-person one, as shown in \ref{3d_plot} (right).  

Figure \ref{3d_exp_ana} (left) displays the comparison of the performances between our model and the baselines in terms of the evaluation metric (Equation \ref{eq:evmet}). The Transporter network is trained on observations in third person and as can be seen, correctly extracts the keypoints on the \emph{pixel plane}. As discussed in Section \ref{sec:relwork}, such a plane differs from $\Sint$ by an unknown projective (and thus non-isometric) transformation. This means that despite the successful keypoint extraction, the geometry of the state space is not preserved, which is reflected by the high error on the plot. This is a general limitation of vision-based approaches: they are unable to recover the intrinsic geometry due to perspective in the case of a three-dimensional scene. Differently from that, our framework extracts an isometric representation and achieves low error independently from the type of observations.

\subsection{Control Task}
In our last experiment we showcase the benefits of our representations in solving downstream control tasks. The motivation is that a geometric and low-dimensional representation improves efficiency and generalization compared to solving the task directly from observations. To this end we design a control task for the Soccer environment consisting in kicking the ball \emph{into the goal}. The reward is given by the negative distance between the (barycenter of the) ball and the (barycenter of the) goal. Observations are views in third person perspective. In each episode the agent and the ball are initially placed in a random location while the ball is placed in the center. The maximum episode length is $20$ steps. 

We train a number of models via the popular reinforcement learning method \emph{Proximal Policy Optimization} (PPO; \cite{schulman2017proximal}). One model (\emph{End-to-End}) receives raw observations as inputs. The others operate on pre-trained representations $\mathcal{Z}$ given by the Transporter network, the VAE and our method respectively. All the models implement a comparable architecture for a fair comparison.   

\begin{figure}[th!]
\centering
\includegraphics[width=.7\linewidth]{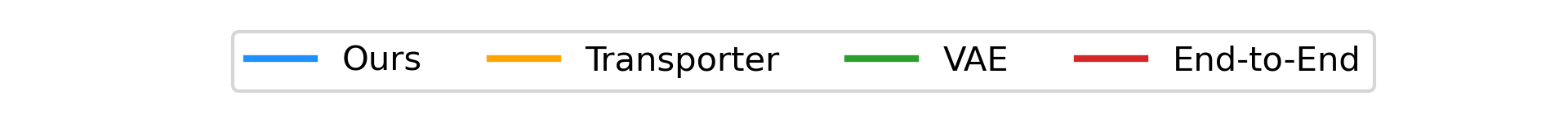}

\begin{subfigure}[b]{.4\linewidth}
        \centering
         \includegraphics[width=\linewidth]{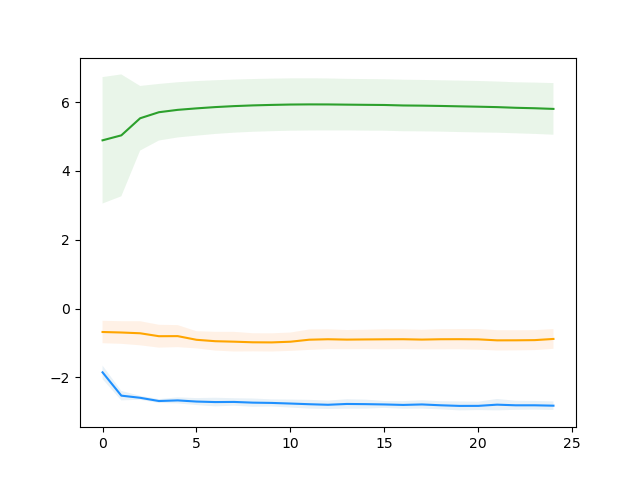}
\end{subfigure}
\hspace{2em}
\begin{subfigure}[b]{.4\linewidth}
        \centering
      \includegraphics[width=\linewidth]{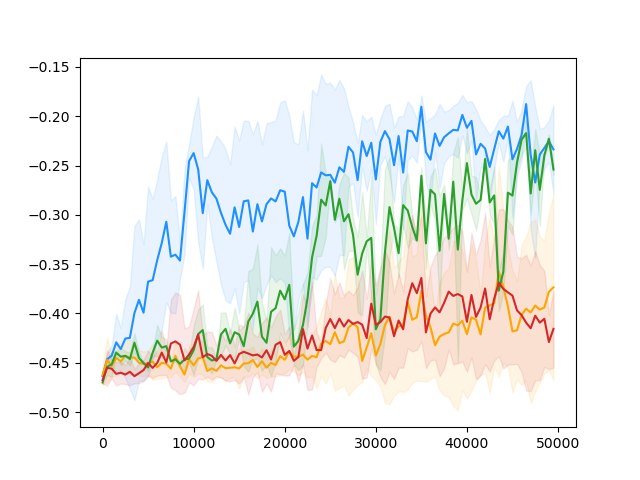}
\end{subfigure}

\begin{picture}(0,0)
\put(-92,8){\scriptsize Epochs}
\put(-150,50){\rotatebox{90}{\scriptsize $\log \mathcal{L}_{\textnormal{test}}$}}
\put(72,8){\scriptsize Steps}
\put(10,50){\rotatebox{90}{\scriptsize Reward}}
\end{picture}
\caption{\textbf{Left}: log-scale plot of the evaluation metric as the training progresses for the Soccer experiment. Observations are in third person. \textbf{Right}: plot of the reward gained via reinforcement learning on top of different representations.}
\label{3d_exp_ana}
\end{figure}
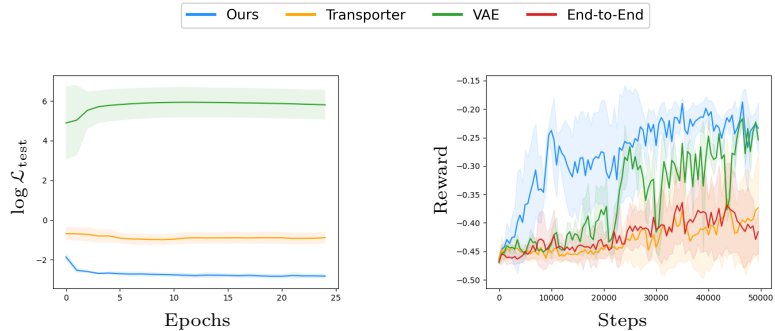

Figure \ref{3d_exp_ana} (right) displays the reward gained on test episodic runs as the training by reinforcement learning progresses. As can be seen, our geometric representation enables to solve the task more efficiently than both the competing representations (Transporter and VAE) and the end-to-end model. Note that the Transporter not only does not preserve the geometry of the state space, but has the additional disadvantage that the keypoint corresponding to the agent and the object can get swapped in the output of $\varphi$. This causes indeterminacy in the representation and has a negative impact on solving the task. Due to this, the Transporter performs similarly to the end-to-end model and is outperformed by the generic and non-geometric representation given by the VAE. In conclusion, the results show that a downstream learner can significantly benefit from geometric representations of observations in order to solve downstream control tasks. 

\section{Conclusions and Future Work}\label{sec:concfut}
In this work we proposed a novel framework for learning representations of both an agent and an object the agent interacts with. We designed a system of losses based on a theoretical principle that guarantees isometric representations independently from the nature of observations and relying on supervision from performed actions alone. We empirically investigated our framework on multiple scenarios showcasing advantages over computer vision approaches.  

Throughout the work we assumed that the agent interacts with a single object. An interesting line of future investigation is extending the framework to take multiple objects into account. In the stochastic context (see Section \ref{sec:stocha}) an option is to model $\zext$ via multi-modal densities, with each mode corresponding to an object. As an additional line for future investigation, our framework can be extended to actions beyond translations in Euclidean space. Lie groups other than $\mathbb{R}^n$ often arise in practice. For example, if the agent is able to rotate its body then (a factor of) the space of actions has to contain the group of rotations $\textnormal{SO}(n)$, $n=2,3$. Thus, a framework where actions (and consequently states) are represented in general Lie groups defines a useful and interesting extension.

\section*{Acknowledgements}
This work was supported by the Swedish Research Council, the Knut and Alice Wallenberg Foundation, the European Research Council (ERC-BIRD-884807) and the European Horizon 2020 CANOPIES project. Hang Yin would like to acknolwedge the support by the Pioneer Centre for AI, DNRF grant number P1.

\section*{Ethical Statement}
We believe that the present work does not raise specific ethical concerns. Generally speaking, however, any system endowing artificial agents with intelligent behavior may be misused e.g., for military applications. Since we propose a representation learning method enabling an agent to locate objects in an environment, this can be potentially embedded into intelligent harmful systems and deployed for unethical applications.   

%
% ---- Bibliography ----
%
% BibTeX users should specify bibliography style 'splncs04'.
% References will then be sorted and formatted in the correct style.
%
\bibliographystyle{splncs04}
\bibliography{main}
\newpage
\section{Appendix}
\subsection{Proofs of Theoretical Results}

\begin{theorem}
Suppose that the representation $\varphi: \ \mathcal{O} \rightarrow \mathcal{Z}$ satisfies: 
\begin{enumerate}
 \item $\varphi$ is equivariant (Definition \ref{defequiv}),
 \item $\varphi$ is injective, 
 \item  for all $o \in \mathcal{O}$ and $a \in \mathcal{A}$ it holds that $\zextp \not = \zext$ if and only if $\zext \in \lfloor \zint, \zint + a \rfloor$ where $(\zint, \zext) = \varphi(o)$ and $(\zintp, \zextp) = \varphi(a \cdot o)$. 
 \end{enumerate}
 Then $\varphi \circ \omega$ is a translation i.e., there is a constant vector $h \in \mathbb{R}^n$ such that for all $s \in \mathcal{S}$ it holds that $\varphi(\omega(s)) = s + h$. In particular, $\varphi$ is an isometry w.r.t. the Euclidean metric on both $\mathcal{S}$ and $\mathcal{Z}$.
\end{theorem}

\begin{figure}[th!]
\centering
\begin{subfigure}[b]{.24\linewidth}
        \centering
         \includegraphics[width=\linewidth]{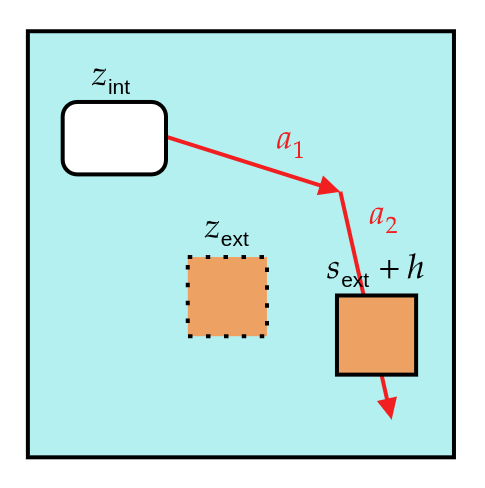}
\end{subfigure}
\hspace{7em}
\begin{subfigure}[b]{.24\linewidth}
        \centering
      \includegraphics[width=\linewidth]{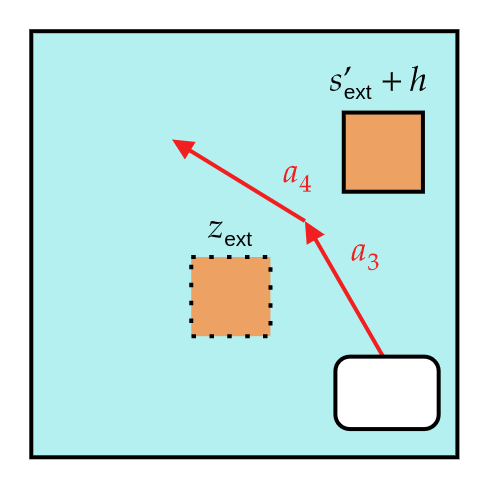}
\end{subfigure}
\caption{Graphical depiction of the proof of Theorem \ref{thminter}. }
\label{figproof}
\end{figure}

\begin{proof}
Pick an arbitrary state $s^0 \in \mathcal{S}$ together with its represented internal state ${z^{0}_{\textnormal{int}}}$ and set $h = {z^{0}_{\textnormal{int}}} - {s^{0}_{\textnormal{int}}}$. For any state $s$, consider  the action $a = \sint - {s^{0}_{\textnormal{int}}}$. Equivariance then implies that $\zint = {z^{0}_{\textnormal{int}}}+ a = \sint + h$. This shows that the claim holds for internal states. 

To prove that the same happens for external states, suppose by contradiction that there is a state $s$ such that $\zext \not = \sext + h$. Consider any path during which the agent interacts with the object without passing through $\zext$. Formally, this means considering a sequence of actions $ a_1, \cdots , a_r$ such that (see Figure \ref{figproof}, left): 
\begin{itemize}
\item $\zext$ and $\sext + h$ do not belong to $\lfloor \zint + a_1 + \cdots + a_{i-1}, \zint + a_1 + \cdots + a_i \rfloor $ for every $i = 1, \cdots , r-1$,
\item $\zext$ does not belong to $\lfloor \zint + a_1 + \cdots + a_{r-1}, \zint + a_1 + \cdots + a_r \rfloor$ but $\sext +h$ does. 
\end{itemize}
The existence of such a path follows from Assumptions \ref{ass:conn} and \ref{ass:rea}. After interaction the state becomes $s' = a_r \cdot (a_{r-1} \cdots (a_1 \cdot s ))$ with $\sextp \not = \sext$ because of Assumption \ref{ass:cont}. One can then consider a path back to the initial agent's position $\zint$ i.e., another sequence of actions $a_{r+1}, \cdots, a_R$ such that (see Figure \ref{figproof}, right):
\begin{itemize}
\item $\sextp + h$ and $\zext$ do not belong to $\lfloor \zint + a_1 + \cdots + a_{i-1}, \zint + a_1 + \cdots + a_i \rfloor$ for every $i=r +1 , \cdots, R$,
% nor to $\lfloor \zint +  a_1 + a_2 + a_3 , \zint + a_1 + a_2 + a_3 + a_4 \rfloor$,
\item $a_1 + \cdots  + a_R =0$. 
\end{itemize}
All the conditions imply together that the representation of the object remains equal to $\zext$ during the execution of the actions $a_1, \cdots , a_R$. Since the actions sum to $0$, the representation of the agent does not change as well. But then $\varphi(\omega(s)) = \varphi(\omega(\sint, \sextp))$ while $\sext \not = \sextp$, contraddicting injectivity. We conclude that $\zext = \sext + h$ and thus $z = s +h$ as desired. 
\end{proof}

\subsection{Pseudocode for Loss Computation}
\begin{algorithm}[ht]
    \caption{Loss Computation}
    \label{algmain}
    \begin{algorithmic}        
    \footnotesize
          \Require Batch $\mathcal{B} \subseteq \mathcal{D}$, models $\varphi_\textnormal{int}, \varphi_\textnormal{ext}, \varphi_\textnormal{cont}$
        \Ensure Loss $\mathcal{L}$
        \State $\mathcal{L} = 0$
        \ForAll{$(o, a, o') \in \mathcal{B}$}  
            \State Compute $\zint = \varphi_\textnormal{int}(o)$, $\zext = \varphi_\textnormal{ext}(o)$, $\zintp = \varphi_\textnormal{int}(o')$, $\zextp = \varphi_\textnormal{ext}(o')$,  $w = \varphi_\textnormal{cont}(o)$,  $w' = \varphi_\textnormal{cont}(o')$
    \EndFor
    \State Compute the classes $C_-, C_+$ via Otsu's algorithm based on $\{ d_\mathcal{W}(w,w')\}$ 
    \ForAll{$(o, a, o') \in \mathcal{B}$}  
        \State Compute $\Lint(o,a,o')$ via Equation \ref{eqlint} 
        \State Compute $A = \{ d_\mathcal{W}(w', w''), \  d(\zintp, \zintp')\}$ \textbf{for} $o''$ marginalized from $\mathcal{B}$
        \State Based on $A$ compute $\Lcont(o,o')$ via Equation \ref{infonce}
        \If{$d_\mathcal{W}(w,w') \in C_-$}
        \State Compute $\Lext(o,a,o') = \mathcal{L}_-(o,a,o')$ via Equation \ref{eqposneg} (left)
        \Else 
        \State Compute $\Lext(o,a,o') = \mathcal{L}_+(o,a,o')$ via Equation \ref{eqposneg} (right)
        \EndIf
        \State $\mathcal{L} \leftarrow \mathcal{L} + \Lint + \Lext + \Lcont$
    \EndFor
    \end{algorithmic}        
\end{algorithm}

\end{document}